# Mobile Exergames: Activity Recognition Based on Smartphone Sensors

David Craveiro[1] and Hugo Silva[1]

[1] Polytechnic of Coimbra, Coimbra Institute of Engineering, Coimbra, Portugal
a2019102658@isec.pt, a21190495@isec.pt

**Abstract.** Smartphone sensors can be extremely useful in providing information on the activities and behaviors of persons. Human activity recognition is increasingly used for games, medical, or surveillance. In this paper, we propose a proof-of-concept 2D endless game called Duck Catch & Fit, which implements a detailed activity recognition system that uses a smartphone accelerometer, gyroscope, and magnetometer sensors. The system applies feature extraction and learning mechanism to detect human activities like staying, side movements, and fake side movements. In addition, a voice recognition system is combined to recognize the word "fire" and raise the game's complexity. The results show that it is possible to use machine learning techniques to recognize human activity with high recognition levels. Also, the combination of movement-based and voice-based integrations contributes to a more immersive gameplay.

## 1 Introduction

The computing and sensing technologies evolution in the recent years made smartphones a common and powerful device. The recent smartphones also incorporate more and better sensors that provide new opportunities to explore the intelligent contexts creation. An Ambient Intelligence system is a digital environment that proactively, but sensibly, support people in their daily lives [1]. As a result, many works on Human Activity Recognition (HAR) field have been done in the recent years.

The opportunities explored in the papers on HAR include entertainment and gaming areas for smartphones. This fact has created a new opportunity for gaming business and the user's engagement has been increasing in the recent years [2]. These days in the gaming industry, there are several games that incorporate physical activity during its gameplay. This kind of games are known as exergames [3].

The consoles brought the exergames to the spotlight, but they require special equipment for this purpose and are not suitable to play outside. Besides, most of mobile exergames do not promote an immersive movement interaction [3]. This is due to lack of realistic mapping between the game activities and the player activities in real life.

Nowadays, humans are less active than before, so it is important to promote human physical activity. According to several studies, exergames help to improve physical activity and people enjoy playing them [4], [5]. However, the possibility to cheat in exergames demotivates the players to interact as intended [5]. So, it is also important to design exergames to ignore or punish the cheating behaviour.

The main goal of this paper is to build a proof-of-concept mobile 2D infinite runner game for Android platform, entitled as Duck Catch & Fit, using an immersive movement interaction. In addition, the game only requires a smartphone to play, where the complexity of the game is achieved by combining movement and voice interaction. The game consists of an actor that must do lateral movements with his body to avoid obstacles, and with his voice to fire and catch the duck.

To achieve the goal, we implement and train a human activity recognition system to identify real and fake side movements by using the smartphone's accelerometer, gyroscope, and magnetometer sensors to collect data and predict in real-time. We also use the Android speech recognizer to predict the word "fire" said by the user.

The results in this paper show that is possible to recognize staying, side movements and fake side movements using Random Forest (RF) algorithm, with an accuracy of 95.10%. The results also show that it is also possible to create a voice recognition system using the Android speech recognition [6].

The main contributions of this paper are the following:

1) Create an available public dataset for staying, side movements, and fake side movements, from accelerometer, gyroscope, and magnetometer smartphone sensors.
2) Explore the combination of feature extraction and selection to improve model performance.
3) Application of state-of-the-art classification Machine Learning (ML) models, and identify the best, for human activity recognition, applied to smartphone accelerometer, gyroscope, and magnetometer sensors data.
4) Make possible and improve the connection between a user and a smartphone game, to promote a better user experience.

The rest of this paper is organized as follows. Section 2 presents related work on mobile exergames and HAR on smartphone and Section 3 describes the system architecture. Section 4 presents the field of study and Section 5 describes the collected data. Section 6 describes the used ML techniques and Section 7 presents the tests made and discuss the results. Section 8 describes future work on this topic and Section 9 presents the conclusions.

# 2 Related Work

This section presents related research papers for mobile exergames and HAR.

## 2.1 Type

The works in [3], [7]–[10] refer to studies using mobile exergames and HAR, the work in [5] refers to study on identifying cheating behaviour of HAR, while the work in [2] studies the level of expertise of the participants on a smartphone game, all of them using ML to recognize and predict, in real time, the activities.

## 2.2 Sensors

The works in [2], [3], [8], [9] used the accelerometer and gyroscope sensors, while work in [7] used the accelerometer, gyroscope and magnetometer sensors. The works in [10], [11] used the accelerometer sensor.

## 2.3 Dataset

Regarding the dataset, in [3] the collection of data was done by six volunteers while doing the activities of standing, move left, move right, squat and jump. Each instance has a window size of 64 values with a 50Hz frequency, which means 1.28 seconds of movement and represents the best duration of the activities to be classified. A continuous activity is the one where the actor is steady without any change to his movement and cyclic. In opposite, a non-continuous activity is the quick movement to avoid collision with an obstacle. For the continuous activities and in test mode, the application applies an overlap of 50% in the collected sliding windows, meaning that new data will come after 32 set of values, and added to the previous 32 set of values. For the non-continuous activities, the collection of data is stopped automatically after the collection of 64 set of values, not occurring any kind of overlap.

In [7] the used dataset was a public available mobile sensors dataset, which includes five daily living activities, namely walking, standing, running, moving upstairs and moving downstairs. Each activity is performed for 5 seconds in multiple attempts, with a frequency rate of 100Hz for the accelerometer and gyroscope, while for magnetometer was 50Hz. The missing samples of the dataset were imputed using extrapolation with the K-Nearest Neighbor (KNN) technique.

In [5] the data was gathered from 12 participants, where the fake activities, as jumping, squatting, stomping, fake jumping, fake squatting, and fake stomping, were performed by the participant on doing physical motions or manipulating the smartphone in their own way to attempt to simulate the activity.

In [2] data was collected from 38 participants, which had a history of playing mobile games, based on the player’s hand movements.

In [8] to collect the data, nine individuals performed six different activities, namely walking, climbing up the stairs, climbing down the stairs, sitting, standing and laying.

Six signals were recorded, three from each dimension of each sensor, with a sampling rate of 50Hz and stored as time series.

In [9] the data was collected from ten volunteers, performing six different activities, namely walking, standing, sitting, lying down, up the stairs, and down the stairs. Each sensor runs at 50Hz to store raw data in units of sample, and each sample includes 128 values.

In [10] data was collected from five users, using smartphones kept in front right pant pocket, doing five static activities, namely sit on floor, sit on chair, lying left, lying right, and lying straight. Regarding the predict of six dynamic activities, they include slow walk, brisk walk, normal walk, jogging, running, climbing upstairs, and climbing downstairs.

In [11] the authors used a benchmark dataset named “HASC2010corpus” [12], which contains 6791 activity files with tri-axial accelerometer data, collected from 540 subjects with a sampling rate of 10-100Hz. The dataset is composed by the following columns: terminal type, frequency, activity, gender, height (cm), weight (kg), shoes type, floor type, place, terminal position on the body and terminal mount. The sequence data includes all the activities, where each activity was continued for 5 seconds or more resulting in a window time of 120 seconds. Then, to intensify the relevant properties of the signal, a window overlap technique was used with 50% overlap in a window with size of 450 instances and duration of 4.5 seconds.

### 2.4 Filter

Regarding the data noise and the used filters, in [7] it was used a low-pass filter to improve the quality of the captured data, due to the best performance in the existing studies.

In [2] to remove noise, the signals were preprocessed using the Savitzky-Golay smoothing filter.

In [8] noise was filtered using median and 20Hz Butterworth filters. A Butterworth filter is a type of signal processing filter designed to have a frequency response as flat as possible in the passband and roll-offs towards zero in the stopband [9]. A second 3Hz Butterworth filter was applied to eliminate effect of gravity in accelerometer signals. After, values were normalized to (-1, 1) interval.

In [10] Butterworth and median filter were used because of noise or abnormal spikes occur in the collected sensory data.

In [11] to separate the gravity from de body acceleration a first order Butterworth low-pass filter was used with a corner frequency of 0.3Hz.

### 2.5 Features

Regarding the features, in [3] the dimensions for each sensor are four, namely x, y, z and magnitude, where magnitude is calculated using the formula in (1).

$$mag = \sqrt{x^2 + y^2 + z^2} \quad (1)$$

The extraction of time-domain features includes the mean, standard deviation, maximum value, minimum value, correlation coefficient between each pair of x, y and z axes, and signal-magnitude area. The frequency domain features include the first 6 FFT coefficient magnitudes, energy and entropy. As the left and right movements were very similar, where the difference is essentially in the accelerometer's x axis, the authors decided to create a subsystem dedicated to the classification of these movements. This subsystem only considers temporal domain features from the accelerometer's x axis, where the extracted features include the x axis raw data, maximum and minimum value, the mean of each quarter of the signal, and the range.

In [7] the features were extracted from two sources, namely raw signal and maximum peaks of the signal. Regarding raw signal, the extracted features include the mean, median, standard deviation, variance, maximum value, and minimum value. For the maximum peaks of the signal features, the mean, standard deviation, variance, median, and five most significant distances of the maximum peaks were extracted.

In [5] the features extracted from the accelerometer were the user acceleration and gravity, while for the gyroscope, rotation rate and attitude were used. The user acceleration, gravity and rotation rate are composed by three values, x, y and z, where attitude contains pitch, roll, and yaw. The selected features were normalized between the values 0 and 1.

In [2] a 22 time domain features were used from the preprocessed data, extracted on each axis of accelerometer and gyroscope, namely maximum amplitude, minimum amplitude, maximum latency, minimum latency, latency to amplitude ratio, absolute latency to, amplitude ratio, peak-to-peak signal value, peak-to-peak time, peak-to-peak slope, mean, standard deviation, skewness, kurtosis, energy, normalized energy, entropy, mean of absolute values of the first difference, mean of absolute values of second difference, mean of absolute values of first difference of normalized signal and mean of absolute values of second difference of normalized signal. A wrapper-based feature selection method was used to identify the most relevant, where it selects a subset of features by optimizing the classifier performance.

In [8] regarding the features, the Euclid magnitudes of the values of three dimensions were calculated and merged into one dataset. The class codes were added at the end, as well as the number of each individual. In the end, the dataset consists of 2947 records with 561 features.

In [9] for the feature extraction, there are three parameters that need to be evaluated and selected, namely the length of the sliding window, the displacement of consecutive windows, namely a 50% overlap of data, and the features parameters. For the created features, they include the difference between the value of two consecutive values in array, magnitude of the three variables x, y, and z, FFT, mean, standard deviation, average of standard deviation, maximum, minimum, signal magnitude area (SMA), energy, and entropy.

In [10] different subset of features was applied, where feature subset 1 contains mean, standard deviation, and variance. The feature subset 2 contains additionally the minimum and maximum features, where for feature subset 3 it is added the entropy. Finally, the feature subset 4 adds a new feature based on jerk (rate of change in acceleration).

In [11] the mean, variance, minimum, maximum, standard deviation, Median Absolute Deviation (MAD), skewness, kurtosis, mean frequency and inter quartile range were calculated. Then to obtain orientation independent sensor information, the jerk of body acceleration was calculated along the three axes and APTD feature was calculated using the formula in (2).

$$APTD = \frac{\sum_{n=0}^{N-1} |A_{fn} - A_{in}|}{N} \quad (2)$$

This feature consists in the sum up of all the amplitude differences from trough to peak and peak to trough, where $A_{fn}$ is the final amplitude, and $A_{in}$ is the initial amplitude. Then the sum is divided by N which is the number of trough to peak and peak to trough pairs. At last, FFT was used to extract frequency domain features, where mean frequency and subsets of various FFT coefficients were used.

### 2.6 Models and Performance

Regarding the ML models and performance, in [3] the best results were obtained with supervised RF algorithm, with an accuracy of 99% with the 10-fold cross validation method.

In [7] for the training and test, 10-fold cross-validation was used, for eight different ML models, namely Decision Tree (DT), KNN, RF, AdaBoost (AB), Logistic Regression (LR), Support Vector Machine (SVM), Naive Bayes (NB), and Neural Network (NN). Over the single accelerometer sensor, the AB algorithm outperformed all other ML algorithms with 100% in all metrics, while SVM algorithm performed worst with 78.48% precision, 77.76% recall, 77.76% accuracy, and 77.83% F1-score. For the accelerometer and magnetometer sensors, RF and AB outperformed all classifiers with 100% in precision, recall, accuracy and F1-score. The overall performance, while using the previous two sensors, improved comparing with the single sensor data. For the accelerometer, magnetometer, and gyroscope sensors, RF and AB outperformed all other ML algorithms with 100% precision, recall, accuracy and F1-score.

In [5] the algorithms used for the training were Multilevel Perceptron Network (MLP) and RF classifier. The data was split into a training and test set in the proportion of 80% and 20%, respectively. The RF classifier was used on the recorded dataset using 10-fold cross validation. The results show that MLP performed well on classifying fake squatting with 100% accuracy and fake stomping with 92.9% accuracy. On the other hand, jumping and squatting were classified with an accuracy of 21.4% and 28.6%, respectively. The RF model classified squatting and fake stomping with accuracy of 95.8% and 94.4%, respectively. The RF model worst performance were jumping with 83.8% accuracy and stomping with 88.1% accuracy. The RF model has a much higher accuracy that the MLP model.

In [2] the performance of the classifiers was evaluated using 10-fold cross-validation for training KNN, RF and NB. The accelerometer sensor classified more accurately with an average accuracy of 85.08% using KNN, while gyroscope achieved the maximum accuracy of 84.21% using RF. The fusion of both sensors improves the performance. The maximum accuracy was achieved for KNN with 92.10%, 89.47% for RF, and 71.92% for NB. The other metrics were also higher using KNN as compared to RF and NB classifiers.

In [8] the training data consisted of 80% of the total dataset, and test data the remaining 20%. The performance of the classifiers was evaluated using 5-fold cross-validation, using DT, SVM, KNN and Ensemble classification models (Boosting, Bagging and Stacking). Regarding DT algorithm, when binary DT is used, the success rate (number of correct classified cases divided by total number of cases) of 53.1% is achieved. When branching limit is raised to 20, the classification success increased to 91.7%. As branching increases, success rate and calculation time also increases. When branching limit is set to 100, success rate increases to 94.4%. For SVM, when a cubic polynomial kernel is used, success rate of 99.4% is achieved. Regarding KNN, for k=1 success rate was 97.1%, while for k=3 success rate was 97.5%. For Ensemble classifiers, AB classifies 97.4% of the records successfully, Bagging has a successful classification ratio of 98.1%, and Stacking, that consists of 30 KNN classifiers, has 98.6% of success.

In [9] the training data consisted of 70% of the total dataset, and test data the remaining 30%. The supervised classifier used was SVM. In order to compare with previous research from Anguita [13] the recognition detection rate (number of correct classified cases divided by total number of cases) obtained was 2752/2957 samples (93.38%), quite higher than the Anguita result of 89%. Using the collected data from this study, the detection rate is above 89%, where the lying activity has low recognition rate and was perceived to act sitting, because of similarity of smartphone status between sitting and lying activities.

In [10] the performance of the classifiers was evaluated using 5-fold cross-validation for DT, SVM linear, SVM quadratic, SVM cubic, KNN, Boosted Ensemble, and Bagged Ensembled. The application of feature subset 1 on the data, implies a best overall accuracy of 90.9%, performing well for static activities with very little misclassification error for classes lying right and lying straight, while for dynamic classes fails in learning to differentiate between them. For feature subset 2, the results are even better for static classes, improving the accuracy for running, jogging, slow and brisk walk, but the accuracy dips a little for normal walk, and for dynamic classes the misclassification is large. There is no improvement in static activity when applying feature subset 3, however, significant relative improvement can be observed in dynamic activity accuracies. When applying feature subset 4, the DT classifier has lower accuracy as they are non-robust and prone to overfitting. The KNN classifier fails as it is highly dependent upon its neighbor. SVM performs poorly with linear kernel, however, its accuracy goes up with increase in polynomial degree of kernel. The prediction accuracy for walk classes increase significantly, as for running and jogging. Climbing upstairs and downstairs have tolerable misclassification error.

In [11] the performance of SVM, Linear Discriminant Analysis (LDA), KNN and RF were compared. The LDA resulted in accuracy of 93.17%, the KNN with k=5 resulted in accuracy of 92.33% and the RF, with 20 trees, resulted in accuracy of 94.17%. The SVM was evaluated with linear kernel resulted in accuracy of 94.33%. The SVM, using a radial basis function (RBF) kernel with C=1 and = 0.01, resulted in accuracy of 96.46%. In order to prevent overfitting, the dataset was split into three parts, 60% for train, 30% for cross validation split in 10 folds, and 10% for test. The activities with less accuracy are walk 87% and upstairs with 91.3%.

## 3 System Architecture

In this section we present the system architecture for the activity recognition. In the general process steps subsection, we have a general vision of the process for the activity recognition. In activities subsection, the activities that we want to detect are described. For the data collection subsection, the several steps of the recognition system are described, from the data collection until activity classification. Finally, in feature extraction, all the extracted features from the data collection are presented.

### 3.1 General process steps

The general process steps for the activity recognition include the data collection step, where the sliding window and the classification frequency are very important, where the first may depend on the type of activity. Also, it is very important to collect data with the less noise possible, so that the model may be the most reliable possible.

The next step is the feature extraction, which is one of the most important, where the raw data is processed to extract relevant features that can improve the classification step. The selection step is used to reduce the previous extracted features through selection algorithms, that can turn the classification process simpler. Finally, the classification step where one instance of the selected features is classified.

### 3.2 Activities

The activities are performed by the actor in the game, in order to avoid the collision with objects. There are two main activities to be recognized, namely standing and lateral movement, in which the last can differ on right and left movement.

Standing is the movement to make the actor continue to run without any change to his movement. It is a cyclic movement and continuous.

Lateral movement is the movement to avoid collision with an obstacle. It is a quick movement and not continuous, that differ on the direction, right or left movement.

Also, two more activities were identified, which are the fake right and fake left movements. The purpose is to make the system more reliable against tries to fake the movements, and to encourage the actor to do the activities correctly. Types of movements that fake the system, are the move of the arm while trying to simulate the activities without moving the body.

### 3.3 Data collection

For the data collection, an application for the training mode was built, as it can be seen in Figure 1. Through the interface, the user can choose the activity that will be executed, start and stop the collection. Also, after the collection it is possible to upload the data into the server.

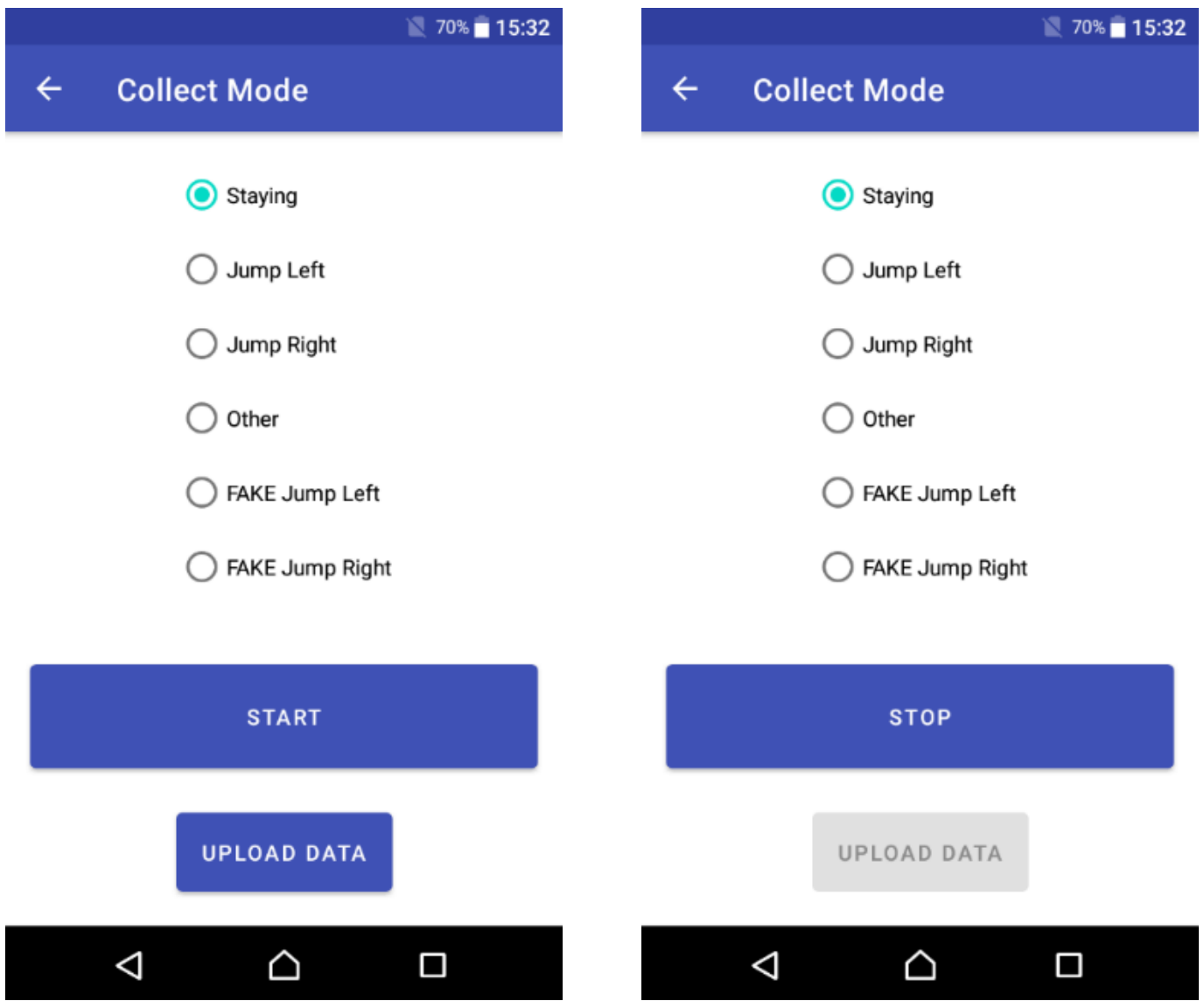

**Fig. 1.** Screenshots of the data collection application for the training.

Figure 2 presents the architecture diagram of the collecting system for the training.

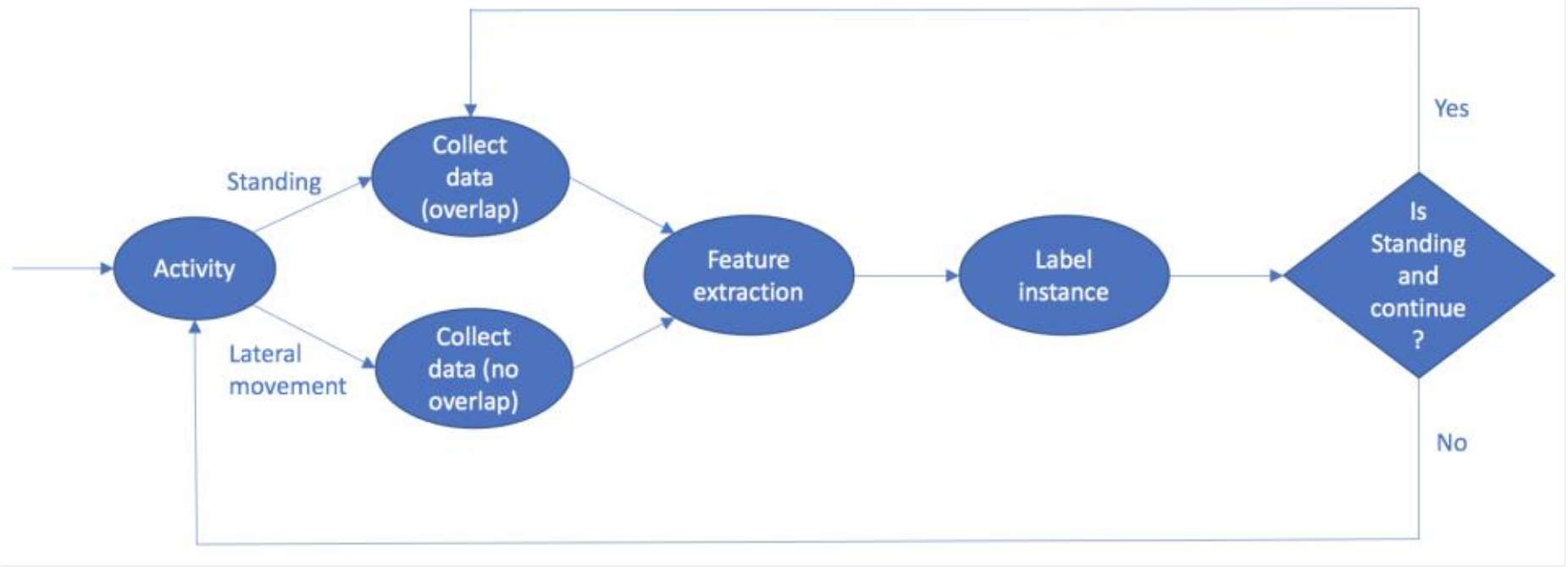

**Fig. 2.** Architecture diagram of the collecting system for the training.

Regarding this diagram of the collecting system, the application gathers set of 64 values (sliding window) from the three axes, x, y, and z, from accelerometer, gyroscope, and magnetometer. The data is adjusted to collect data with a sampling rate of 50Hz, where each instance of the 64 values represents about 1.28 seconds. This rate is due to the SENSOR_DELAY_GAME [14] flag of the Android operating system. Also, this value must be a power of two, because in the extraction phase, it will be used FFT, allowing a faster processing of this algorithm.

Regarding the filters the data collector application use a low pass to remove the gravity from accelerometer data and a high pass filter to remove noise for accelerometer, gyroscope and magnetometer data.

For the continuous standing activity, the application will use an overlap of 50% of the sliding window. This means that new 32 set of data values will be collected and added to the previous 32 values already collected from the previous sliding window. For the lateral movement activities, 64 values for each activity will be collected, where no overlap is applied. The goal of this is to collect data in a cleaner way for short activities and not continuous, like lateral movements.

The user has the possibility of sending the collected instances, to the available server in Comma-Separated Values (CSV) file format.

The execution of the activities during the train considers three kind of user postures for each activity, as it can be seen in Figure 3.

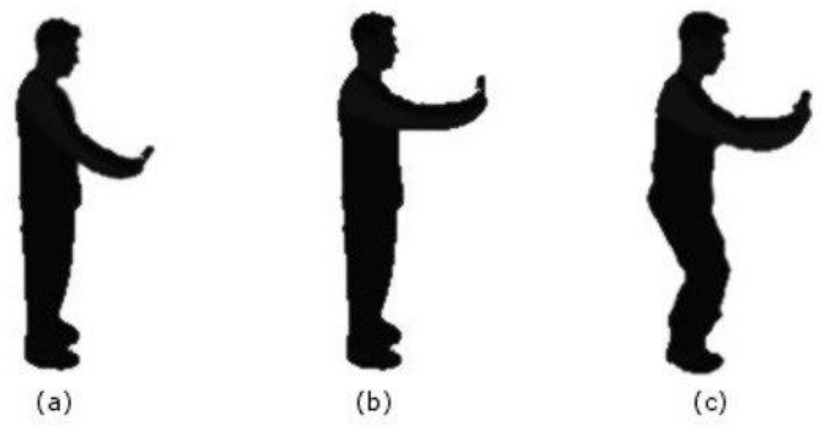

**Fig. 3.** Three postures considered on activity training [3].

In the first position the user is upright with the smartphone down (a). In the second position the user has the same body posture, but now with the device position higher in front of the face (b). Finally, in the third position the movements are performed from a more solid base in which the legs are slightly apart, shoulder-width apart, the legs slightly bent, and the device position raised in from of the face (c). This position should be the recommended posture to realize the movements, although the previous positions should be considered in order to improve the capacity of the system classification.

The diagram regarding the training system is shown in Figure 4.

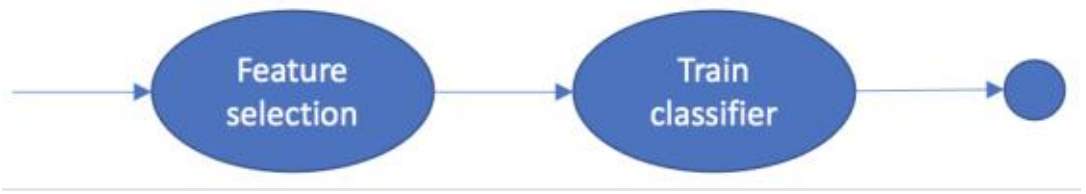

**Fig. 4.** Architecture diagram of the training system.

For the training, the method for selection of the features will be applied to the collected data available in the server, and after the system trains the data regarding the selected classifier

After training, and regarding the data collection test system, an application was built, as it can be seen in Figure 5. This application logs the predicted activity.

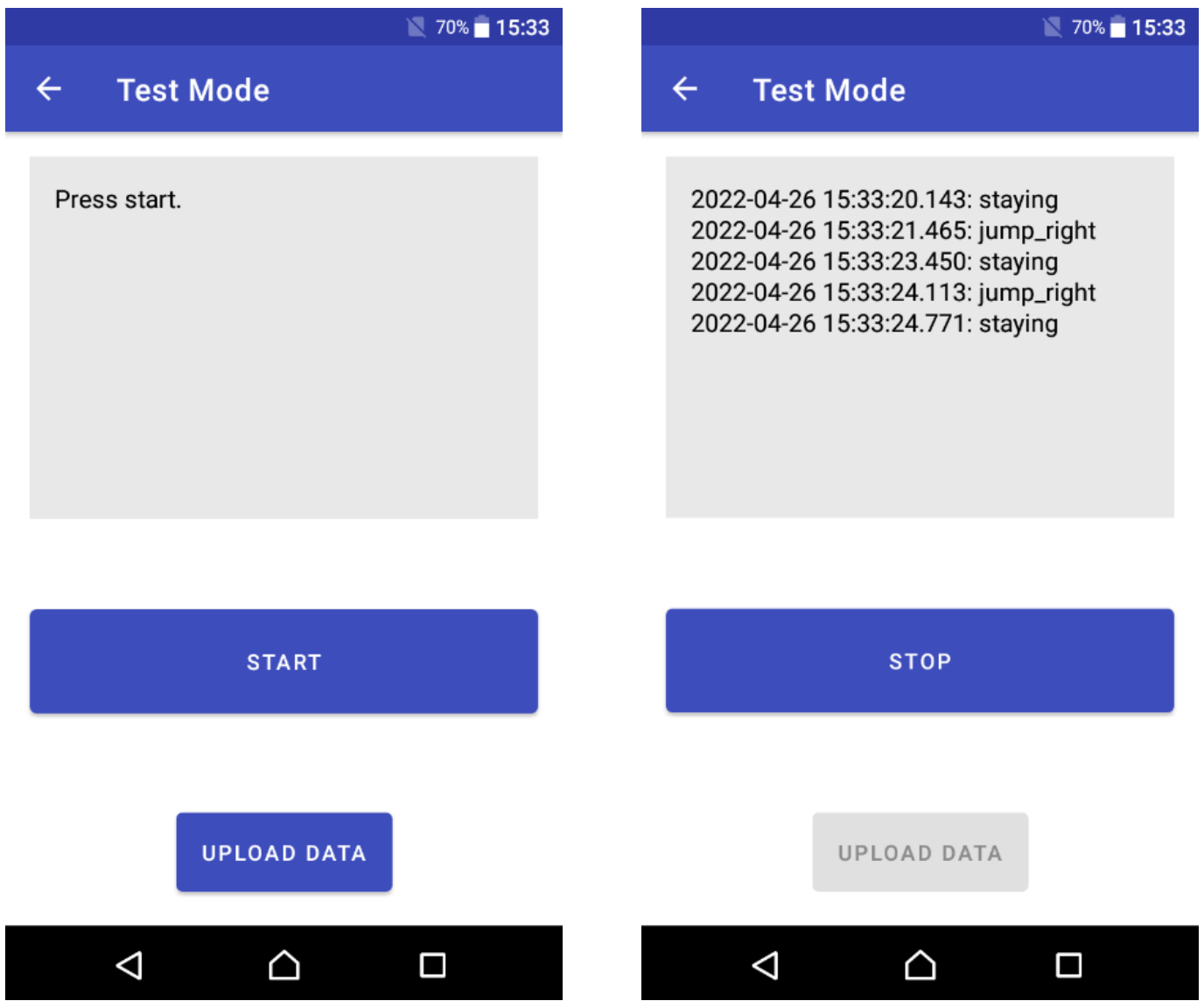

**Fig. 5.** Screenshots of the data collection system for the test mode.

The diagram for the test of the system is shown in Figure 6.

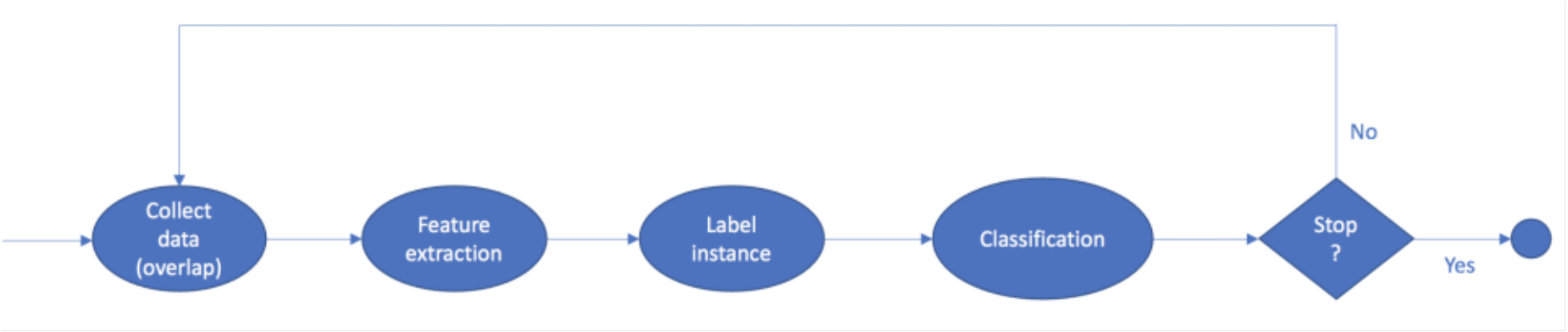

**Fig. 6.** Architecture diagram of the test system.

This diagram provides the workflow for the testing system, where the application is continuously collecting data, with overlap of 50% of the sliding window, and the activity is predicted from the previous trained classifier. This application is helpful in evaluating how good the activities are being predicted, and if more instances of each activity are needed, before using the classifier in the game.

Regarding the voice recognition system, we use the Android Speech Recognizer [6] to collect voice windows in loop and then submit them to Google API. The speech recognition could also work offline if the target language model has been previously downloaded through language and input settings. The game makes use of Google's American English language model to recognize the word fire and then make the character fire a laser. In addition, the partial results mode is enabled in order to get results more often.

**Feature extraction**

In general, the used features are an inspiration of all related works, where they had a positive impact in the ML model performance.

The collected data from the three axes of accelerometer, gyroscope and magnetometer, will be used for the classification of the main activity standing, lateral movement and fake lateral movement. If needed and for performance comparison, for the lateral movement, the detection of right and left movements, the used data will be from the x axis of the accelerometer as in [3].

For the detection of the main activities, temporal domain and frequency domain features will be used. Regarding temporal domain, the used features will be average, standard deviation, variance, minimum value, maximum value, correlation coefficient between the pairs of the x, y and z axes, SMA, and MAD.

The magnitude between the axes x, y and z is calculated using the formula in (1).

The Pearson correlation coefficient between the axes (x, y), (x, z), and (y, z), is obtained by dividing the covariance between the two axes by the product of the standard deviation of the same two axes, in (3).

$$corr(x, y) = \frac{cov(x, y)}{\sigma_x * \sigma_y} \quad (3)$$

The SMA is calculated by summing the magnitudes.

The used features of the frequency domain will be coefficient magnitude of FFT, energy and entropy. The use of FFT will allow to calculate the Discrete Fourier Transform (DFT) of a sequence of values, which converts a signal in the temporal domain to the frequency domain. To be efficient, is necessary that the size of the sequence of values, must be a power of two. The obtained result while using FFT will be two sequences of values, one representing the real part and the second the imaginary part.

Through this process, we obtain the coefficient magnitudes of FFT for each axis of accelerometer, gyroscope, and magnetometer, as well as the coefficient magnitudes of each sensor. The energy is the power of two of the sums of the coefficient magnitudes resulting from FFT. The entropy is calculated through the formula in (4), being useful in differentiating activities with similar energies.

$$H(x) = \sum_{i=1}^{N} p(xi) log_{10} p(xi) \quad (4)$$

Regarding the detection of right and left movements if a single model approach could not be able to differentiate well the two lateral movements, the following features will be used only for the x axis of accelerometer: raw value of x axis, difference between the maximum and minimum value of the position, mean of each quarter, minimum value, and maximum value.

According to [3], the use of the difference between the maximum and minimum value of the position, is because the right movement is characterized by the presence of a maximum preceded by a minimum value, as well as the opposite for the left movement, which can help to distinguish the direction of the movement.

Also, since the signal is characterized by two variations that nullify each other in the global value of the average, it was decided to subdivide in four quarters and calculate for each one its average, increasing the capacity to obtain relevant characteristics to distinguish the variations.

The APTD is calculated for accelerometer and gyroscope axes through the formula (2) and kurtosis is calculated for accelerometer, gyroscope and magnetometer axes [15].

## 4 Field Study

The collection of data was done between April and June of 2022 and will be available for future studies at kaggle.com/datasets/davidcraveiro/duck-catch-and-fit-dataset (Accessed On June 23, 2022).

During the collection period, four participants, including the two authors of this work, were asked to perform all the activities of the game using a smartphone. For that, the volunteers had to agree and sign a Privacy Policy document, which can be found at github.com/djocraveiro/duck-catch-and-fit (Accessed on June 23, 2022). It was informed that the collected data does not contain any information that can identify who did the action and where it was performed, so privacy is completely guaranteed.

The data collection was also supervised, to ensure that the activities were carried out properly, which required some effort from the monitor and the volunteers.

## 5 Data Visualisation

From the collected data, graphs were drawn for a better understanding of the system. Figure 7 represents the accelerometer evolution from the x, y and z axes, during the sliding window of 64 values.

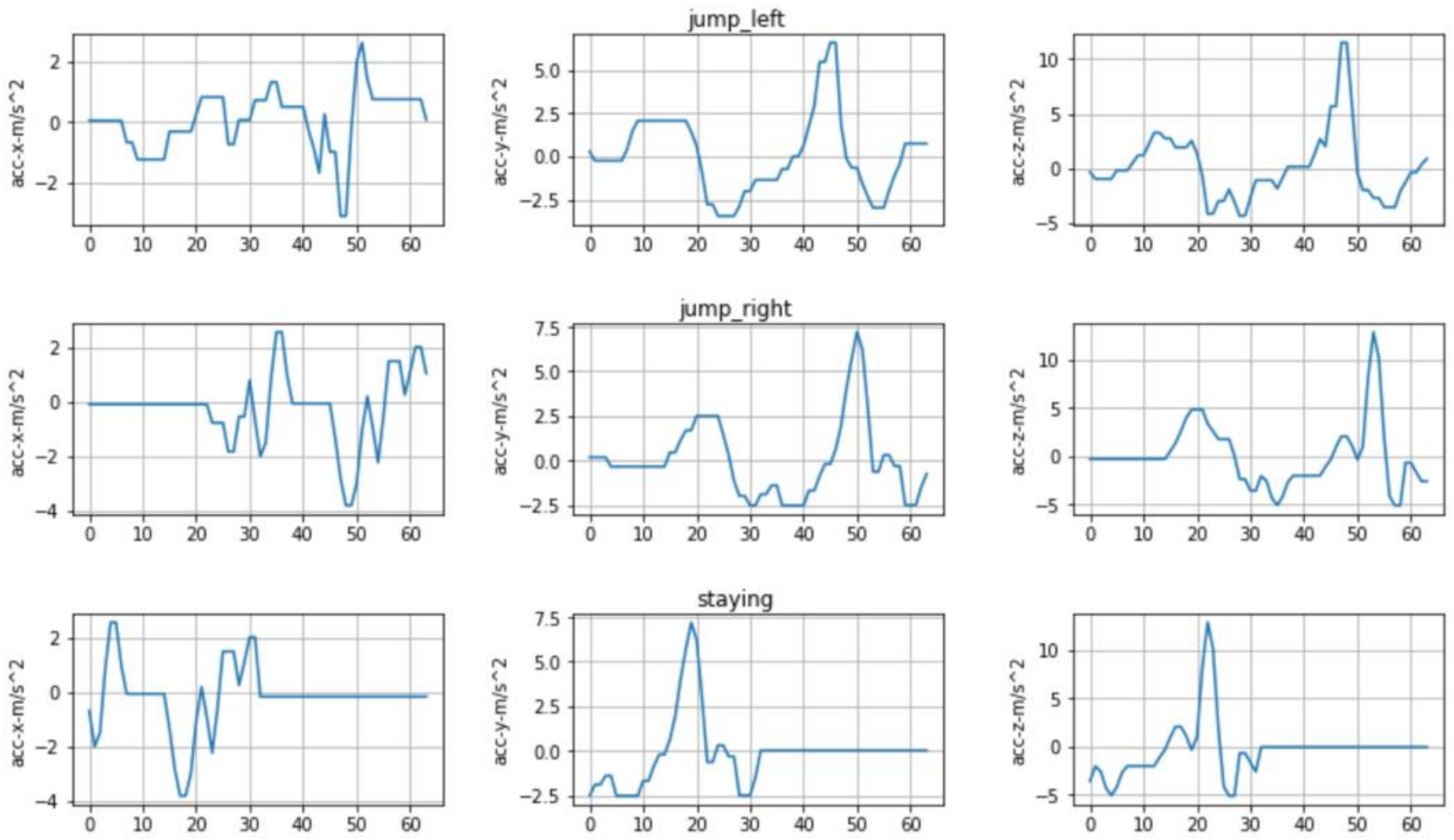

**Fig. 7.** Accelerometer x, y and z dimensions data, during the sliding window of 64 values.

Between the activity jump left and jump right, the data from x, y and z axes of accelerometer, are very similar. Regarding the staying activity, for all the x, y and z dimensions of accelerometer, it seems that there is a signal noise at the beginning, justified by a small movement while pushing the button to record the activity.

Figure 8 shows the gyroscope from the x, y and z dimensions, during the sliding window of 64 values.

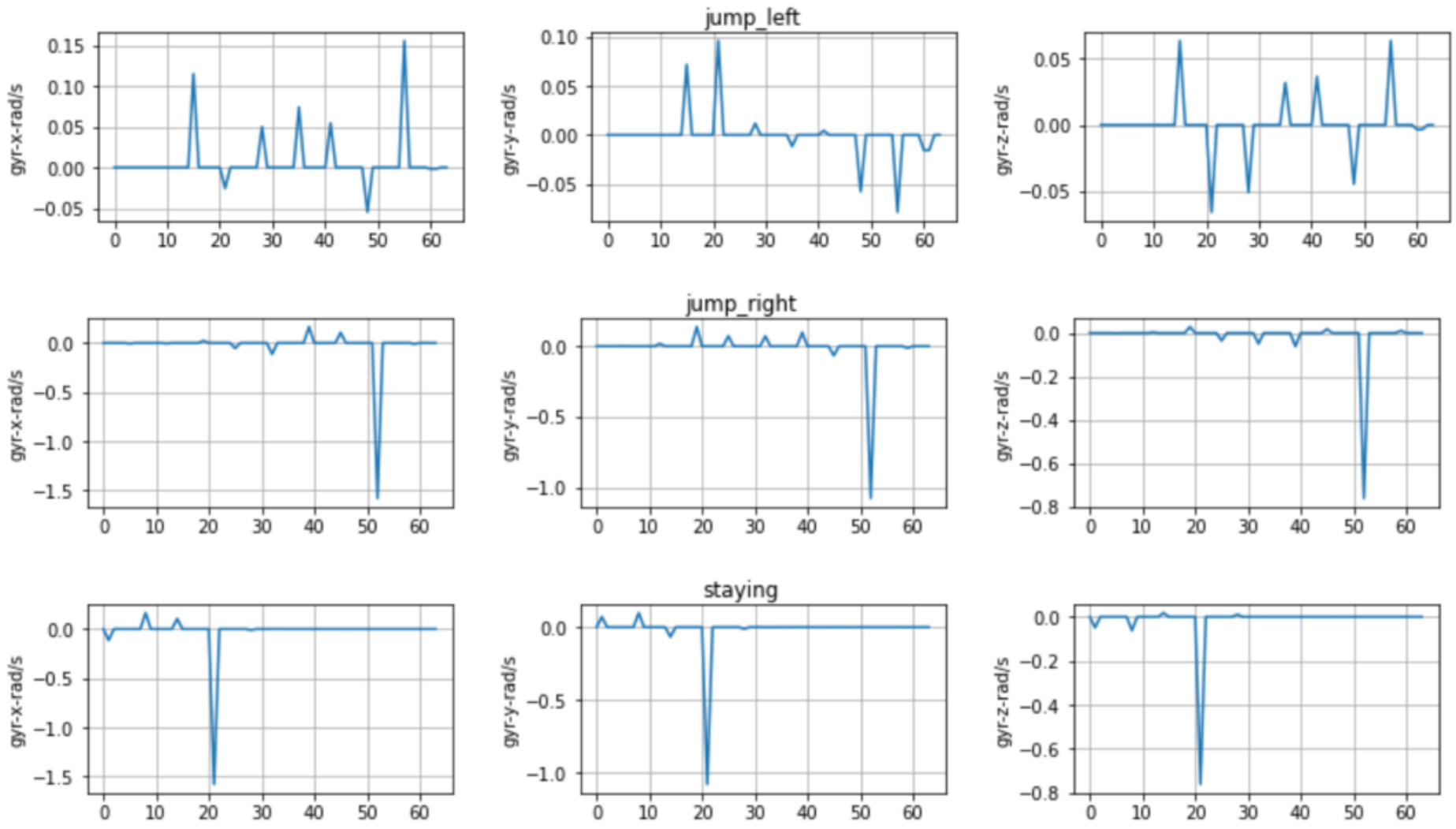

**Fig. 8.** Gyroscope x, y and z dimensions data, during the sliding window of 64 values.

There is not a big variation of values for the gyroscope dimensions, between jump left and jump right activities, except some spikes that occur occasionally.

Figure 9 shows the magnetometer from the x, y and z dimensions, during the sliding window of 64 values.

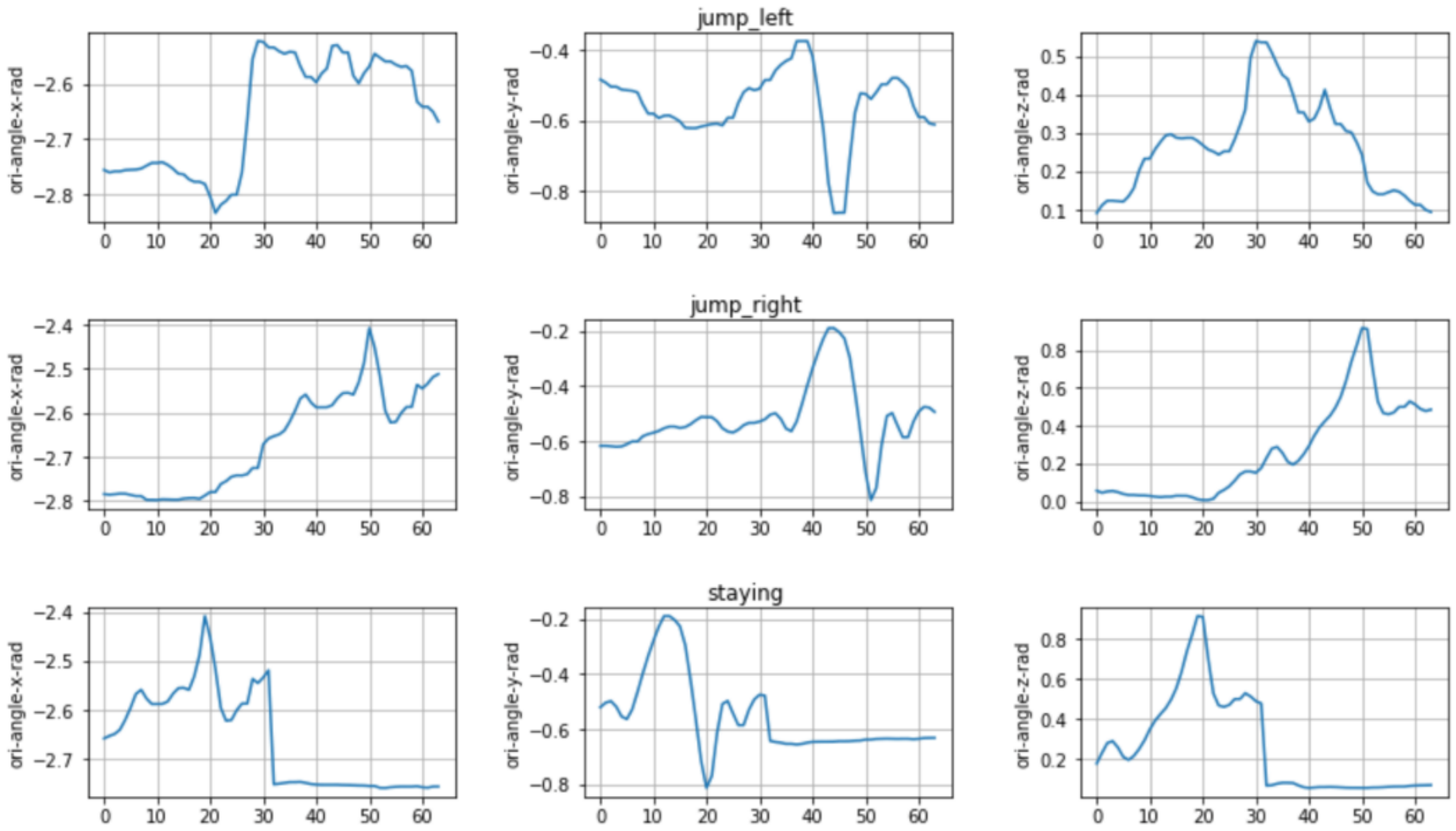

**Fig. 9.** Magnetometer x, y and z dimensions data, during the sliding window of 64 values.

Between the activity jump left and jump right, the trend of data from x, y and z axes of accelerometer, have the same behaviour. Again, for the staying activity it seems that there is a signal noise at the beginning.

Figure 10 represents the box plot graphs for accelerometer x, y and z dimensions, regarding minimum, mean, maximum, variance and standard deviation features.

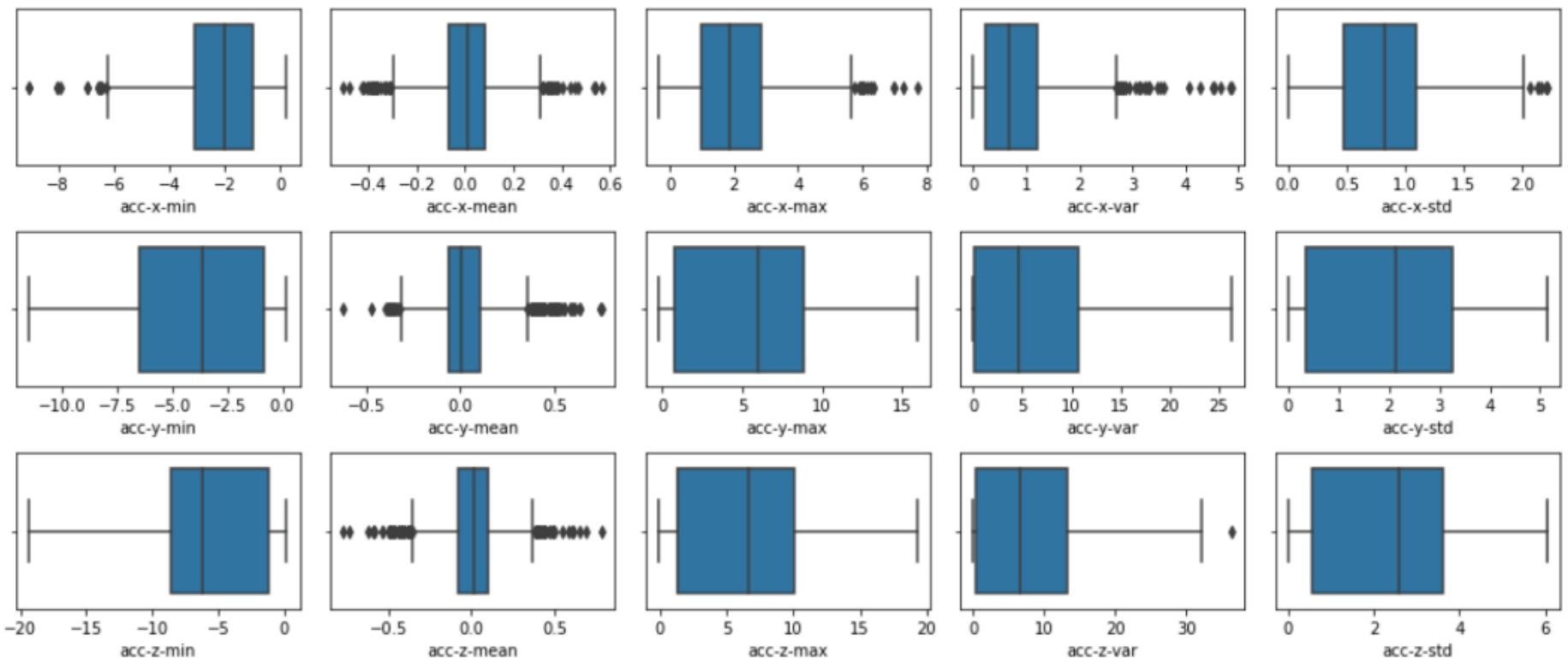

**Fig. 10.** Box plot graphs for accelerometer x, y and z dimensions, regarding minimum, mean, maximum, variance and standard deviation features.

The accelerometer x dimension has a bigger quantity of outliers in comparison with the remaining dimensions. Also, for accelerometer y and z dimensions, mean feature presents a bigger number of outliers, regarding the remaining features.

Figure 11 represents the box plot graphs for gyroscope x, y and z dimensions, regarding minimum, mean, maximum, variance and standard deviation features.

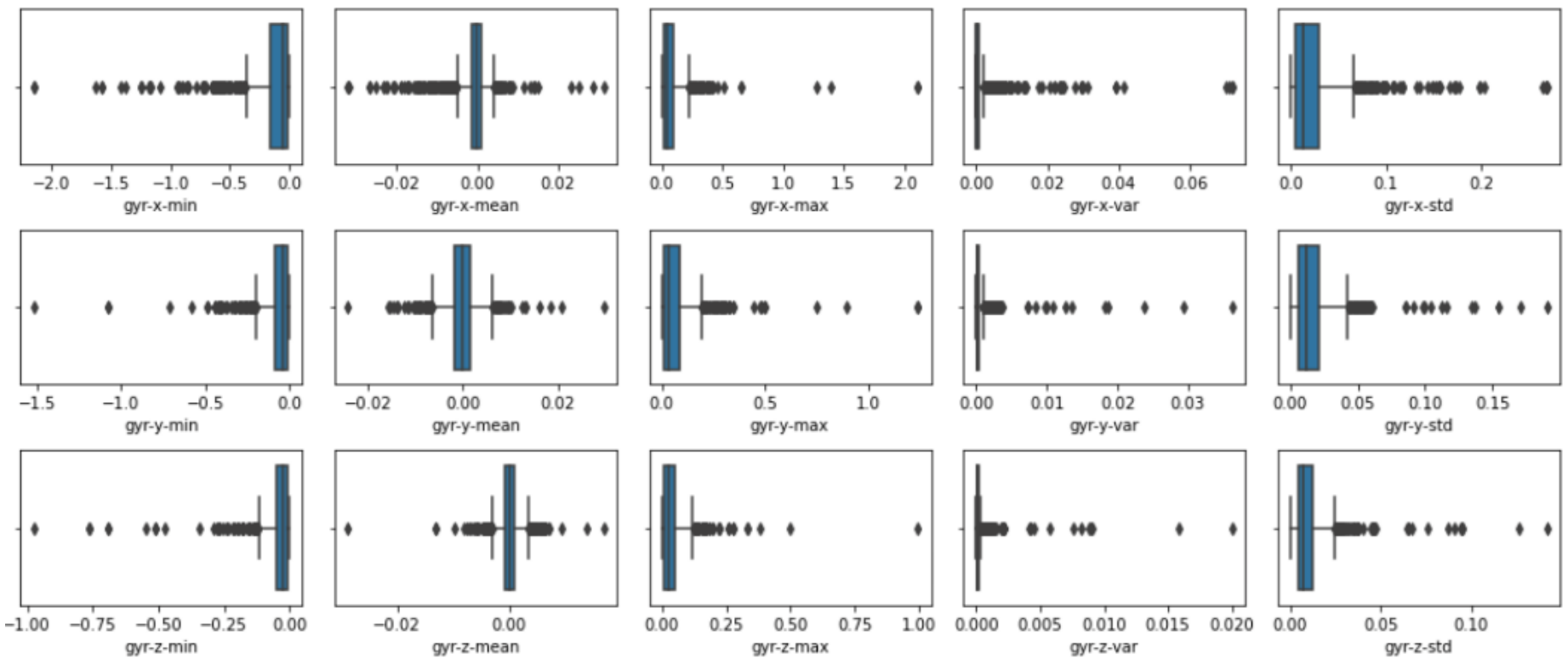

**Fig. 11.** Box plot graphs for gyroscope x, y and z dimensions, regarding minimum, mean, maximum, variance and standard deviation features.

All the features, for all the gyroscope dimensions, present a considerable number of outliers.

Figure 12 represents the box plot graphs for magnetometer x, y and z dimensions, regarding minimum, mean, maximum, variance and standard deviation features.

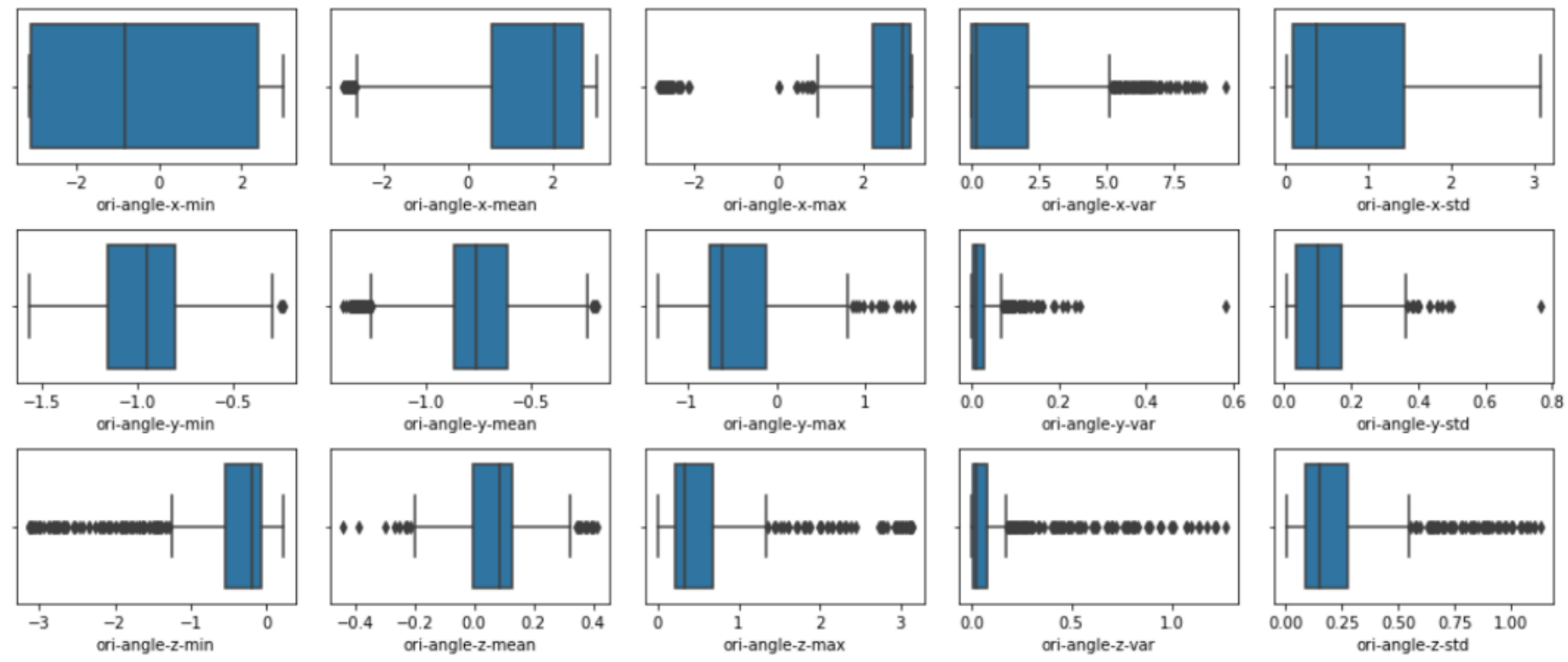

**Fig. 12.** Box plot graphs for magnetometer x, y and z dimensions, regarding minimum, mean, maximum, variance and standard deviation features.

The magnetometer z dimension presents bigger outlier's values for all the features, compared to the remaining feature dimensions.

Figure 13 represents an example of a jump left accelerometer x dimension values, and the application of median and mean filters with different kernel size values.

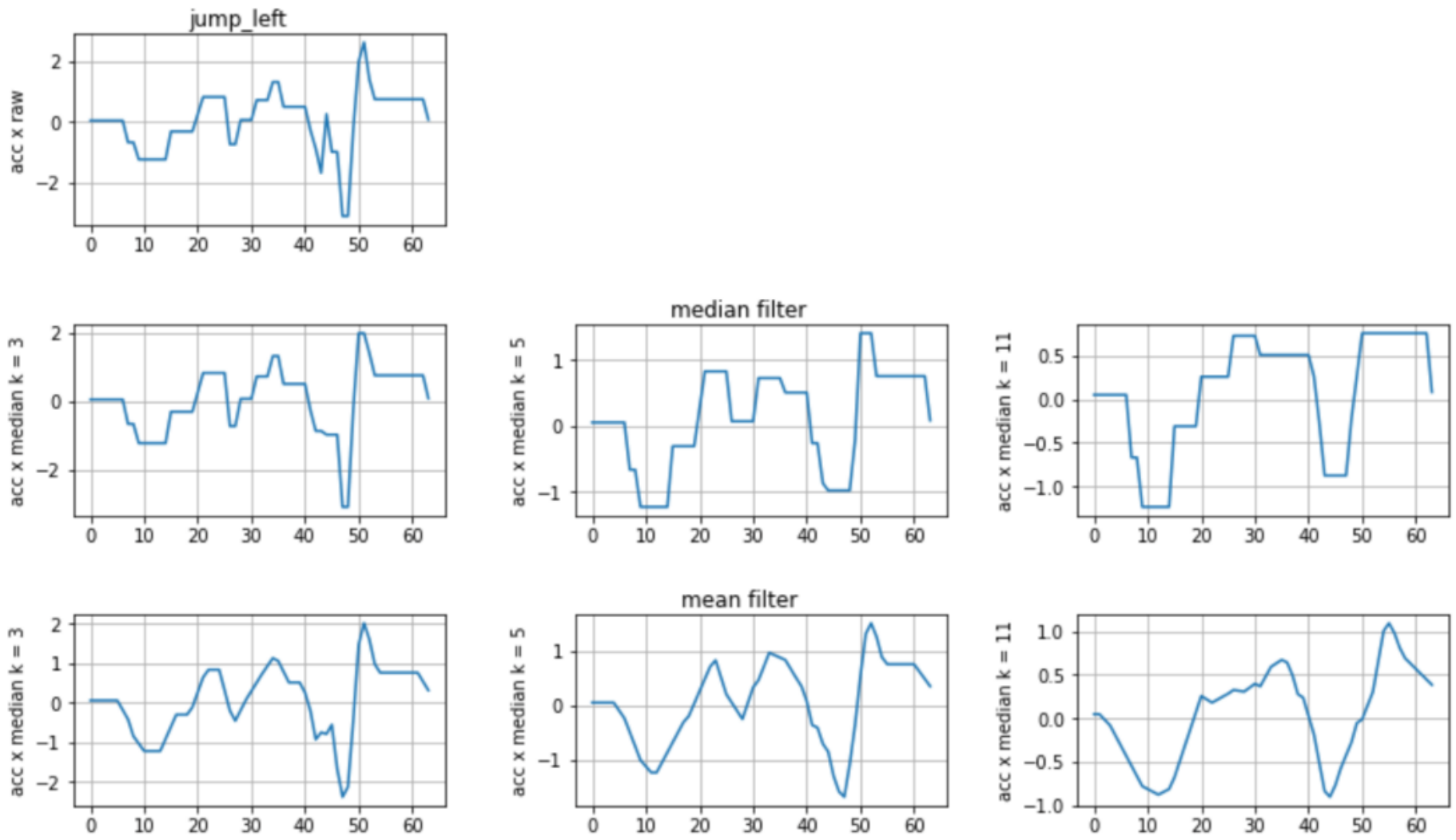

**Fig. 13.** Jump left accelerometer x dimension values, with the application of median and mean filters for different kernel size values.

The application of increasing kernel size values, translates in a smoother signal, as expected. Also, between median and mean filters, median filter aggregates neighbors as a constant value.

Figure 14 shows the distribution of the instances regarding the staying, fake move, jump left and jump right activities.

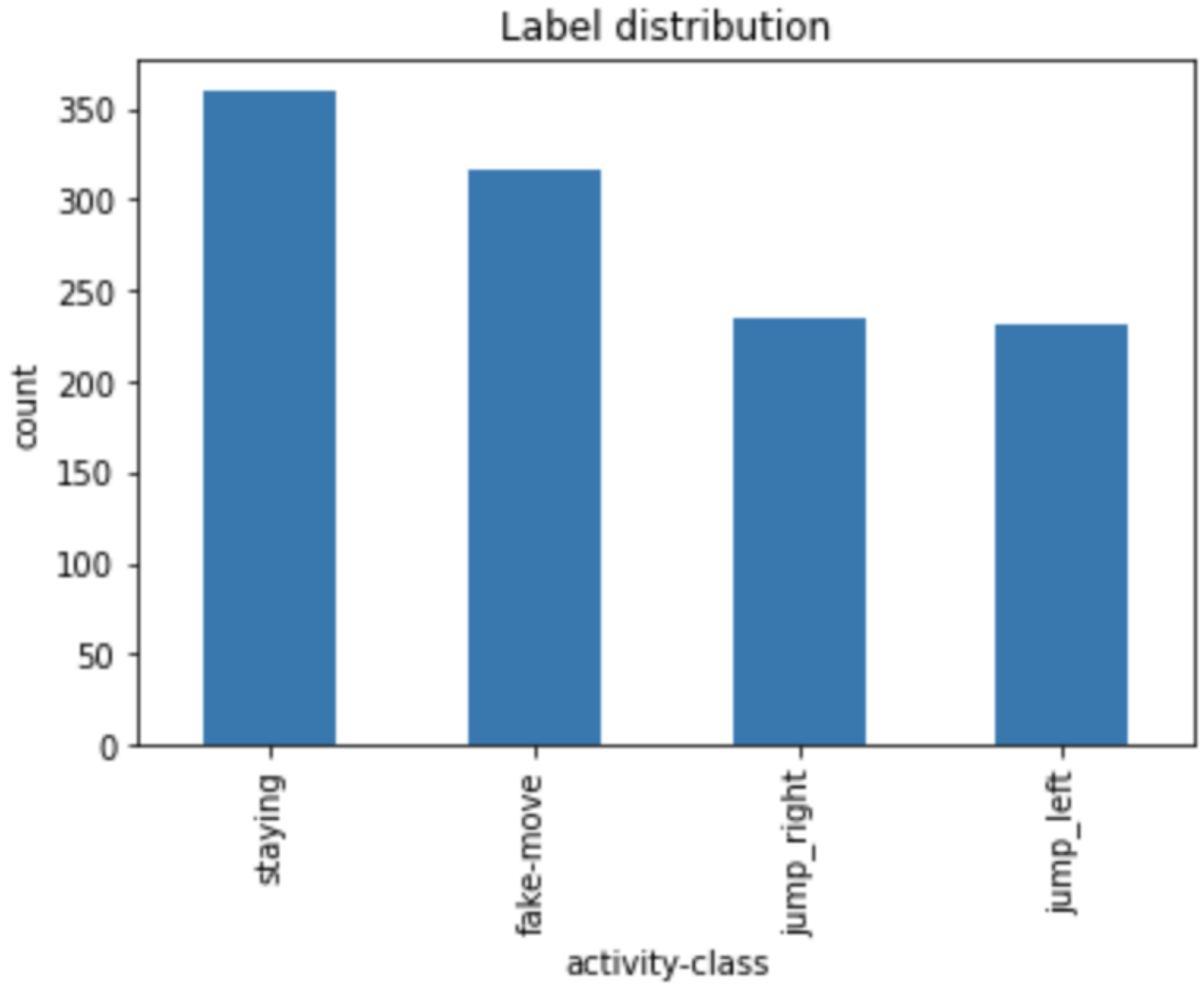

**Fig. 14.** Distribution of staying, fake move, jump right and jump left activities.

The dataset is quite balanced between all the activities, even thought, staying and fake move activities present a bigger number of instances, above 300.

Figures 15, 16 and 17, show the correlation matrixes between accelerometer, gyroscope and magnetometer features, for values greater than 0.3 or lesser than –0.3.

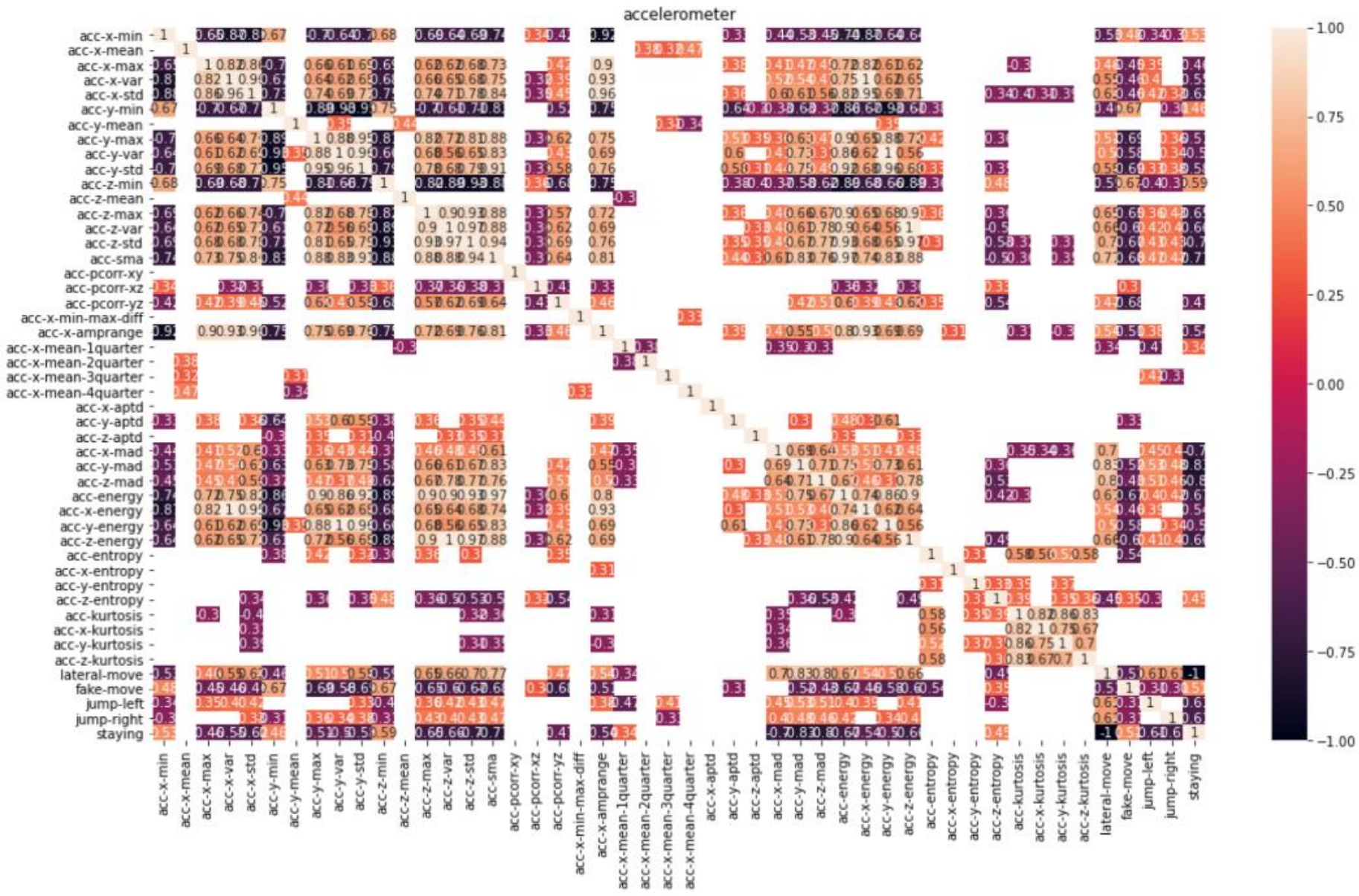

**Fig. 15.** Correlation matrix for accelerometer features.

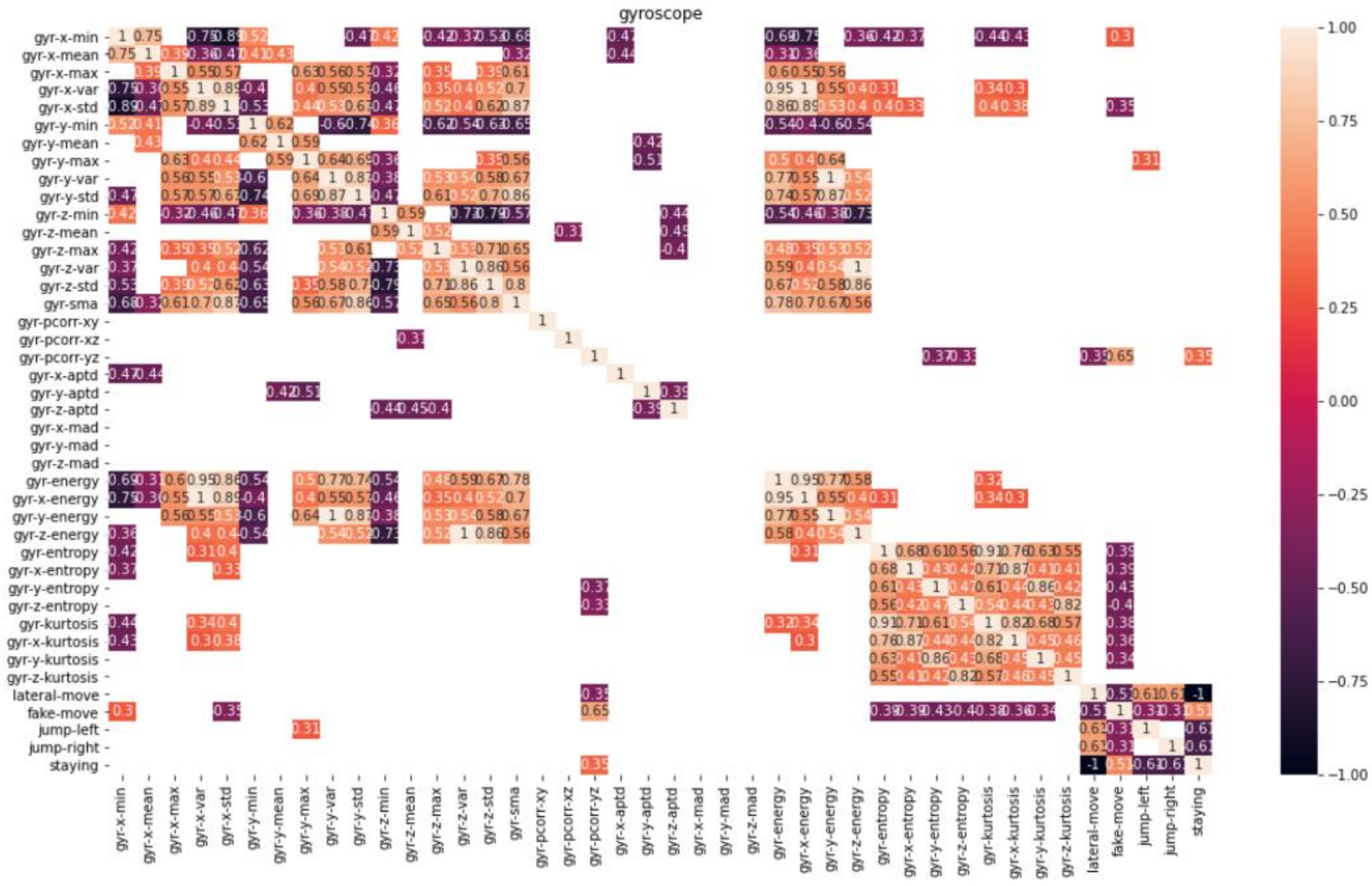

**Fig. 16.** Correlation matrix for gyroscope features.

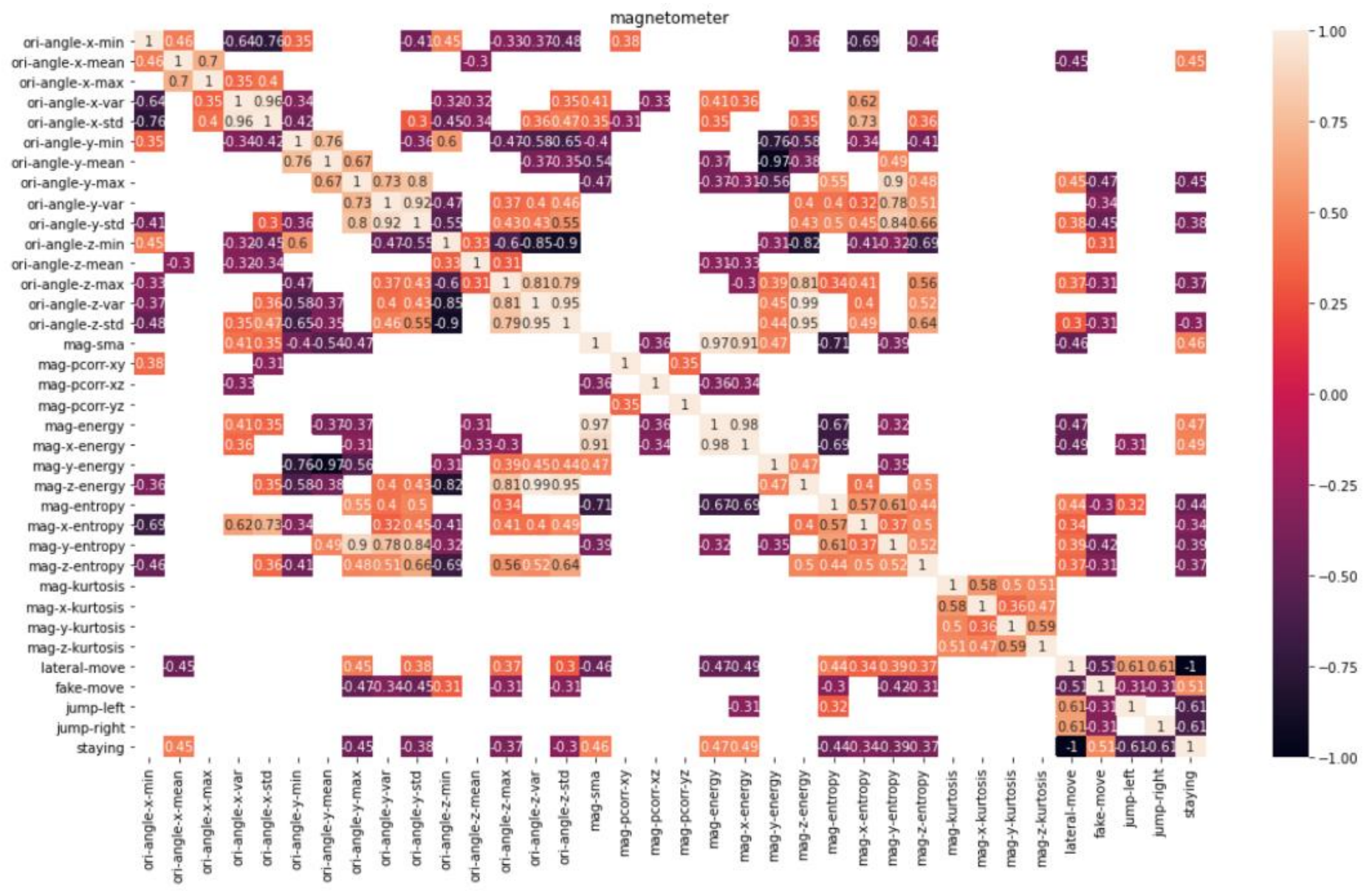

**Fig. 17.** Correlation matrix for magnetometer features.

There is a bigger correlation between features from accelerometer, followed by gyroscope and then magnetometer. In general, there is a considerable correlation between features, and the use of these in the ML models, could lead to good metric results.

# 6 Machine Learning

The five most used algorithms found in related works are RF, KNN, SVM, DT and AB. In addition, the algorithms that shown better classification performance are SVM, RF, AB and KNN.

The type of the used ML algorithms is of classification type, provided by Weka [16]. All of them are trained and tested using the 10-fold cross validation method, also available in Weka. We also use low-pass, median and mean filters to remove the noise from the sensors data, and to be able to improve the classification metrics.

We describe the used algorithms applied to our dataset, inspired by the algorithms used in related works.

DT can categorize or make predictions on how a previous set of questions were answered. It resembles a tree wherein the base of the tree is the root node, from which we obtain a series of decision nodes that depict the decisions to be made. From the decision nodes representing the question or split point, we have leaf nodes which represent the consequences of those decisions or the answers.

RF algorithm uses n decision trees with k aleatory characteristics. The classification is gotten by the vote of all trees, winning the most voted class. While DT consider all possible features splits, RF only select a subset of those features.

KNN is another algorithm which classifies an instance based on the classification of k similar instances. It is called lazy learning algorithm because it does not perform any training when we supply the training data, but simply stores them during the training time and no calculations are made.

SVM is an algorithm which can separate classes through a hyperplane in an N-dimensional space with N as the number of features which distinctly classifies the data points.

AB algorithm, where first we built a model on the training dataset, then a second model is built to rectify the errors present in the first model. This procedure is continued until and unless the errors are minimized.

NB is based on Bayes theorem, where we can find the probability of A happening, given that B has occurred. The assumption made is that the predictors or features are independent, meaning that the presence of one feature does not affect the other. NB classifier is a simple model that describes a particular class of Bayesian network, where all the features are class-conditionally independent.

Simple Logistic Regression (SLR) is a statistical analysis method to predict a binary outcome, such as yes or no, based on prior observations of a dataset. This model predicts a dependent data variable by analyzing the relationship between one or more existing independent variables.

Ensemble is a ML model that combines the predictions from multiple models.

Bagging is a class of ensemble learning method that involves fitting many decision trees on different samples of the same dataset and averaging the predictions.

In the following tables, we can find the results obtained during the test phase of applying the different ML models, using 10-fold cross-validation.

Table 1 and Table 2, show the used instances and ML model performance, using 103 attributes (listed in Appendix A), for the test 1 and 2, and test 1, respectively.

**Table 1.** Used instances for the tests 1 and 2.

| Instance | Value |
|---|---|
| Total | 951 |
| Jump_left | 206 |
| Jump_right | 209 |
| Staying | 287 |
| Fake_move | 249 |

**Table 2.** ML model performance for the test 1.

| Model | Details | Accuracy (%) | Precision (%) | Recall (%) | F1-Score (%) |
|---|---|---|---|---|---|
| Bagging | Random Forest N_Features = 10 | 96.74 | 96.7 | 96.7 | 96.7 |
| Random Forest | N_Features = 10 | 96.63 | 96.6 | 96.6 | 96.6 |
| AdaBoost | Random Forest N_Features = 10 | 96.63 | 96.6 | 96.6 | 96.6 |
| Simple Logistic | | 95.05 | 95.1 | 95.1 | 95.1 |
| SMO (Weka SVM) | RBFKernel + filter type = “Standardize training data” | 94.84 | 94.9 | 94.8 | 94.8 |
| SMO (Weka SVM) | Normalized PolyKernel + filter type = “Standardize training data” | 94.84 | 94.9 | 94.8 | 94.9 |
| SMO (Weka SVM) | PolyKernel | 94.53 | 94.5 | 94.5 | 94.5 |
| Ensemble (Weka vote) | Random Forest N_Features = 10 + Simple Logistic + SMO PolyKernel | 90.85 | 90.9 | 90.9 | 90.8 |
| BayesNet | | 89.80 | 89.5 | 89.8 | 89.7 |
| Decision Tree | Weka REPTree | 89.16 | 89.3 | 89.2 | 89.2 |
| KNN (Weka IBK) | k=30 | 84.33 | 85.1 | 84.3 | 84.4 |
| Naïve Bayes | | 83.38 | 84.1 | 83.4 | 83.1 |

Table 3 shows the ML model performance, using 94 attributes, for the test 2 using the instances from Table 1.

**Table 3.** ML performance for the test 2.

| Model | Details | Accuracy (%) | Precision (%) | Recall (%) | F1-Score (%) |
|---|---|---|---|---|---|
| Random Forest | N_Features = 10 | 96.74 | 96.8 | 96.7 | 96.7 |
| Bagging | RandomForest N_Features = 10 | 96.63 | 96.6 | 96.6 | 96.6 |
| Ensemble | Random Forest N_Features = 10 + Simple Logistic + SMO PolyKernel | 96.31 | 96.3 | 96.3 | 96.3 |
| Ensemble | Random Forest N_Features = 10 + Simple Logistic + SMO RBFKernel + filter type = “Standardize training data” | 96.10 | 96.1 | 96.1 | 96.1 |
| AdaBoost | Random Forest N_Features = 10 | 96.00 | 96.0 | 96.0 | 96.0 |
| SMO (Weka SVM) | RBFKernel + filter type = “Standardize training data” | 95.16 | 95.2 | 95.2 | 95.2 |
| SMO (Weka SVM) | Normalized PolyKernel + filter type = “Standardize training data” | 95.05 | 95.1 | 95.1 | 95.1 |
| Simple Logistic | | 94.74 | 94.7 | 94.7 | 94.7 |
| SMO (Weka SVM) | PolyKernel | 94.53 | 94.5 | 94.5 | 94.5 |
| BayesNet | | 89.48 | 90.2 | 89.5 | 89.5 |
| Decision Tree | | 89.37 | 89.5 | 89.4 | 89.4 |
| KNN | k = 30 | 84.33 | 85.0 | 84.3 | 84.4 |
| Naïve Bayes | | 82.64 | 93.4 | 92.6 | 92.4 |

Table 4 and 5, show the used instances and ML model performance, using 94 attributes, for the test 3.

**Table 4.** Used instances for the test 3.

| Instance | Value |
|---|---|
| Total | 980 |
| Jump_left | 216 |
| Jump_right | 218 |
| Staying | 293 |
| Fake_move | 253 |

**Table 5.** ML performance for the test 3.

| Model | Details | Accuracy (%) | Precision (%) | Recall (%) | F1-Score (%) |
|---|---|---|---|---|---|
| Ensemble (Weka vote) | N_Features = 10 + Simple Logistic + SMO RBFKernel + filter type = "Standardize training data" | 95.20 | 95.2 | 95.2 | 95.2 |
| Random Forest | Random Forest N_Features = 10 | 95.10 | 95.1 | 95.1 | 95.1 |
| AdaBoost | RandomForest N_Features = 10 | 94.79 | 94.8 | 94.8 | 94.8 |
| Bagging | RandomForest N_Features = 10 | 94.69 | 94.7 | 94.7 | 94.7 |
| Ensemble | Random Forest N_Features = 10 + Simple Logistic + SMO PolyKernel | 94.69 | 94.7 | 94.7 | 94.7 |
| SMO (Weka SVM) | RBFKernel + filter type = "Standardize training data" | 93.77 | 93.8 | 93.8 | 93.8 |
| SMO (Weka SVM) | PolyKernel | 93.67 | 93.7 | 93.7 | 93.7 |
| Simple Logistic | | 93.36 | 93.4 | 93.4 | 93.4 |
| SMO (Weka SVM) | Normalized PolyKernel + filter type = "Standardize training data" | 93.06 | 93.1 | 93.1 | 93.1 |
| Decision Tree | | 86.42 | 86.4 | 86.4 | 86.4 |
| BayesNet | | 87.04 | 88.1 | 87.0 | 87.0 |
| KNN | k = 30 | 83.26 | 84.1 | 83.3 | 83.2 |
| Naïve Bayes | | 82.95 | 83.5 | 83.0 | 82.6 |

## 7 Tests and Discussion

The results of the tests from the previous section, show that for test 1, using all features, Bagging algorithm provided the best metric results, with accuracy of 96.74%, precision of 96.70%, recall of 96.40%, and F1-score of 96.70%. We have calculated the RF number of features parameter through the root square of the total number of attributes in the dataset. For the KNN algorithm we have calculated the k value through the root square of the total number of instances in the dataset.

For the test 2, without the following features, acc-x-aptd, acc-y-aptd, acc-z-aptd, gyr-x-aptd, gyr-y-aptd, gyr-z-aptd, gyr-x-mad, gyr-y-mad, and gyr-z-mad, the RF algorithm presented the best metric results, with accuracy of 96.74%, precision of 96.80%, recall of 96.70%, and F1-score of 96.70%. To remove these features, we've analysed the correlation coefficients with the label, and we've also ran the Info Gain attribute evaluator in Weka with Ranker search method.

We have also tried to optimize the RF parameter number of iterations, using the Elbow method and found the value of 129 iterations with accuracy of 96.95%, precision of 97.0%, recall of 97.0% and F1-score of 96.9%. However, when we tried to use the model in the test application, the performance was worse than the model from test 2, so we concluded that the model was overfitting, and we kept the default parameter value.

Finally, for test 3, we have increased the number of instances of the dataset, and we ran the same conditions as test 2. There was one Ensemble model that presented the best metric results, regarding RF, with an accuracy difference of 0.1%. But as we need fast predictions for our game implementation, we have chosen the RF model, which presented an accuracy of 95.10%, precision of 95.10%, recall of 95.10%, and F1-score 95.10%. Also, regarding test 2 and 3, we made experiences with RF using the data collector, with 20 lateral moves, being the model of test 3 the one with less prediction errors.

We have tried the approach used in [3] using one model for the main activity and other for the left and right jumps. For the main activity (lateral-move, staying and fake-move) the RF presented an accuracy of 96.02%, precision of 96.00%, recall of 96.00% and F1-score of 96.00%. For the left and right jumps the RF model presented an accuracy of 95.10%, precision of 95.20%, recall of 95.20% and F1-score of 95.20%. However, the model for the left and right jumps presented greater mean absolute error (MAE) and root mean square error (RMSE) percentages than the model from test 3. We've also made some experiments using these models in the test application, and they performed poorly recognizing the left and right jumps than model of test 3.

We have also tried to remove the outliers, however, did not translate in better results, because with that, we were losing too much information, and our dataset did not have so many instances.

Regarding mean and median filters, they were experimented in order to improve the model performance. However, by applying mean and median filter with k = 11, the RF model presented poor accuracy than RF model of test 3, with an accuracy of 94.79%, precision of 94.80%, recall of 94.80% and F1-score of 94.80%.

A comparison was made between root mean square (RMS) based FFT features, and axes based FFT features. For RF using RMS based FFT features, we obtained a total of 76 attributes and the model presented an accuracy of 94.48%, precision of 94.50%, recall of 94.50% and F1-score of 94.50%. So, the RF model of test 3, that uses axes based FFT features, presented better performance metrics.

During the tests, we realized that test 3 RF model was predicting correctly the first prediction after the jump, although there was a second prediction after the jump that was very few times wrongly predicted. This problem might be related to the overlapping window technique. This issue will cause the game character to move left or right after a jump. In order to prevent this behaviour, we have ignored the following predictions in a time window of 1s after a first prediction of jump left or jump right activities.

Overall, we felt happy with the RF model from test 3, as when applied to our test application and game, presented very good results on predicting the movements, as well as detecting very well the cheat movements.

Regarding the voice interactions, it works well most of the times, but we have noticed that the voice recognition system showed a little delay after we speak the word "fire". Besides that, we did not use window overlap in the voice recognition system, so if the word got split in two windows the voice recognition system does not work properly.

The game, where screenshots can be seen in Figure 18, was tested with the volunteers using the same smartphones used for data collection.

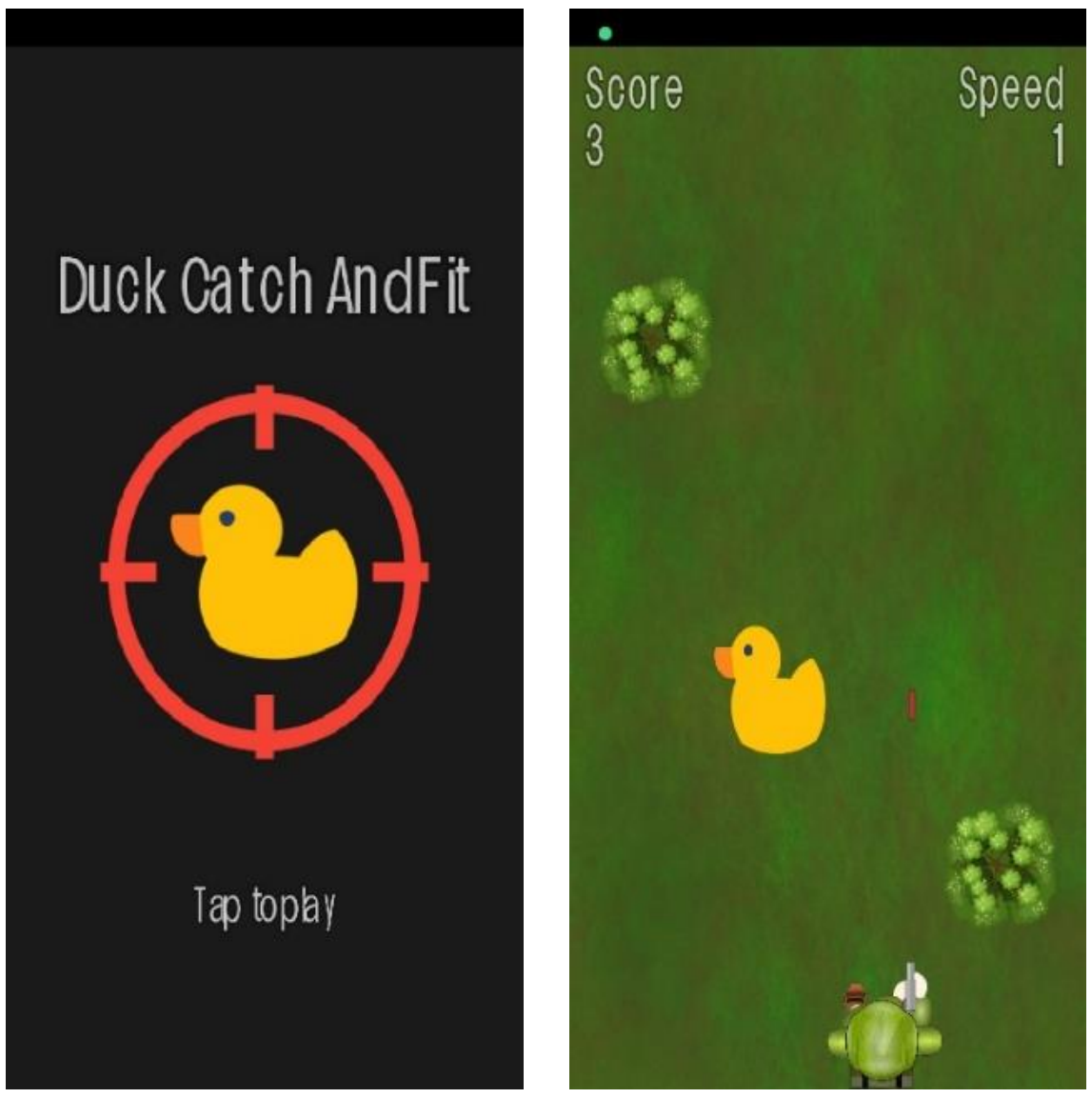

**Fig. 18.** Screenshots of the Duck Catch & Fit game.

We observed that cheating behaviour detection worked most of the time. The volunteers praised that they were more involved into the game, and the speed increase, makes the game more challenging over time. However, they stated that voice recognition has a noticeable delay. In addition, we observed the promotion of physical activity was fulfilled because after a few games the volunteers started to sweat.

## 8 Future Work

As future work we intend to adapt the game difficulty level by increasing the speed adjusted to the player's experience based on accelerometer, gyroscope and magnetometer data.

Further, future tests and data collection must be conducted, including more volunteers to increase the dataset in order to improve the performance of the ML models. In addition, also the Butterworth filter could be applied to our dataset, in order to evaluate the impact on the model's performance.

## 9 Conclusions

HAR in our daily living scenarios brings positive outcomes for novel applications to promote public health. The use of smartphone inertial sensors, as accelerometer, gyroscope, and magnetometer, are the most used to capture and analyse data from daily living activities, because they do not require hardware device setup.

During this work, we developed a system for real-time body movement and voice recognition, applied to a mobile game, to promote the practice of physical activity, without the normal touch interaction. We used several ML algorithms provided by Weka, to recognize lateral side movements and avoid cheating. The used dataset was created by the authors and volunteers, and made available for the public. The experimental results revealed that the RF algorithm demonstrated the best performance, combining the accuracy, precision, recall and F1-score of 95.10%, and the speed of prediction. For the recognition system, we used the Android speech recognition submitted to Google API. However, sometimes we observe that there is a recognition delay.

The results were very satisfactory, where the participants felt motivated and involved with the game itself, besides the physical activity it provides. The participants also praised about how well the system responded and the robustness against cheat attempts.

## References

[1] C. Ramos, "Ambient intelligence - A state of the art from artificial intelligence perspective," in *Lecture Notes in Computer Science (including subseries Lecture Notes in Artificial Intelligence and Lecture Notes in Bioinformatics)*, 2007, vol. 4874 LNAI, pp. 285–295, doi: 10.1007/978-3-540-77002-2_24.

[2] M. Ehatisham-ul-Haq, A. Arsalan, A. Raheel, and S. M. Anwar, "Expert-novice classification of mobile game player using smartphone inertial sensors," *Expert Systems with Applications*, vol. 174, p. 114700, Jul. 2021, doi:

10.1016/J.ESWA.2021.114700. [Online]. Available: https://doi.org/10.1016/j.eswa.2021.114700. [Accessed: Mar. 14, 2022]

[3] A. Almeida and A. Alves, "Activity recognition for movement-based interaction in mobile games," in *Proceedings of the 19th International Conference on Human-Computer Interaction with Mobile Devices and Services, MobileHCI 2017*, Sep. 2017, pp. 1–8, doi: 10.1145/3098279.3125443 [Online]. Available: https://dl.acm.org/doi/10.1145/3098279.3125443. [Accessed: Mar. 02, 2021]

[4] K. Norozi, R. Haworth, A. A. Dempsey, K. Endres, and L. Altamirano-Diaz, "Are Active Video Games Effective at Eliciting Moderate-Intensity Physical Activity in Children, and Do They Enjoy Playing Them?," *CJC Open*, vol. 2, no. 6, pp. 555–562, Nov. 2020, doi: 10.1016/j.cjco.2020.07.006. [Online]. Available: https://pubmed.ncbi.nlm.nih.gov/33305216/. [Accessed: Apr. 16, 2022]

[5] E. Kock, Y. Sarwari, N. Russo, and M. Johnsson, "Identifying cheating behaviour with machine learning," in *33rd Workshop of the Swedish Artificial Intelligence Society, SAIS 2021*, 2021, doi: 10.1109/SAIS53221.2021.9484044 [Online]. Available: https://ieeexplore.ieee.org/document/9484044. [Accessed: Mar. 15, 2022]

[6] Android Developers, "Speech Recognizer," *Android Developers*, 2022. [Online]. Available: https://developer.android.com/reference/android/speech/SpeechRecognizer. [Accessed: Jun. 23, 2022]

[7] I. M. Pires, F. Hussain, G. Marques, and N. M. Garcia, "Comparison of machine learning techniques for the identification of human activities from inertial sensors available in a mobile device after the application of data imputation techniques," *Computers in Biology and Medicine*, vol. 135, Aug. 2021, doi: 10.1016/j.compbiomed.2021.104638. [Online]. Available: https://pubmed.ncbi.nlm.nih.gov/34256257/. [Accessed: Mar. 15, 2022]

[8] E. Bulbul, A. Cetin, and I. A. Dogru, "Human Activity Recognition Using Smartphones," in *ISMSIT 2018 - 2nd International Symposium on Multidisciplinary Studies and Innovative Technologies, Proceedings*, 2018, doi: 10.1109/ISMSIT.2018.8567275 [Online]. Available: https://ieeexplore.ieee.org/abstract/document/8567275. [Accessed: Apr. 21, 2022]

[9] D. N. Tran and D. D. Phan, "Human Activities Recognition in Android Smartphone Using Support Vector Machine," in *Proceedings - International Conference on Intelligent Systems, Modelling and Simulation, ISMS*, Jul. 2016, vol. 0, pp. 64–68, doi: 10.1109/ISMS.2016.51 [Online]. Available: https://ieeexplore.ieee.org/document/7877190/. [Accessed: Apr. 21, 2022]

[10] I. Roychowdhury, J. Saha, and C. Chowdhury, "Detailed Activity Recognition with Smartphones," in *Proceedings of 5th International Conference on Emerging Applications of Information Technology, EAIT 2018*, 2018, doi: 10.1109/EAIT.2018.8470425 [Online]. Available: https://ieeexplore.ieee.org/abstract/document/8470425. [Accessed: Apr. 21, 2022]

[11] M. Ahmed, A. das Antar, and M. A. R. Ahad, "An approach to classify human activities in real-time from smartphone sensor data," in *2019 Joint 8th International Conference on Informatics, Electronics and Vision, ICIEV 2019 and 3rd*

*International Conference on Imaging, Vision and Pattern Recognition, icIVPR 2019 with International Conference on Activity and Behavior Computing, ABC 2019*, 2019, doi: 10.1109/ICIEV.2019.8858582 [Online]. Available: https://ieeexplore.ieee.org/document/8858582. [Accessed: Apr. 16, 2022]

[12] N. Kawaguchi *et al.*, "HASC challenge: Gathering large scale human activity corpus for the real-world activity understandings," in *ACM International Conference Proceeding Series*, 2011, doi: 10.1145/1959826.1959853.

[13] D. Anguita, A. Ghio, L. Oneto, X. Parra, and J. L. Reyes-Ortiz, "Human activity recognition on smartphones using a multiclass hardware-friendly support vector machine," in *Lecture Notes in Computer Science (including subseries Lecture Notes in Artificial Intelligence and Lecture Notes in Bioinformatics)*, 2012, vol. 7657 LNCS, doi: 10.1007/978-3-642-35395-6_30.

[14] Android Developers, "Sensor Manager," *Android Developers*, Mar. 17, 2022. [Online]. Available: https://developer.android.com/reference/android/hardware/SensorManager. [Accessed: Apr. 19, 2022]

[15] S. Gawali, "Shape of data: Skewness and Kurtosis," *Analytics Vidhya*, May 02, 2021. [Online]. Available: https://www.analyticsvidhya.com/blog/2021/05/shape-of-data-skewness-and-kurtosis/. [Accessed: Jun. 28, 2022]

[16] M. Hall, G. Holmes, B. Pfahringer, P. Reutemann, E. Frank, and I. H. Witten, "The WEKA data mining software: An update Encyclopedia of Machine Learning and Data Mining View project TOETOE Technology for Open English-Toying with Open E-resources [ˈtɔɪtɔɪ] View project The WEKA Data Mining Software: An Update," 2014 [Online]. Available: https://www.researchgate.net/publication/221900777

# Appendices

**Appendix A** - List of all 103 attributes

| Accelerometer | Gyroscope | Magnetometer |
|---|---|---|
| acc-x-min | gyr-x-min | ori-angle-x-min |
| acc-x-mean | gyr-x-mean | ori-angle-x-mean |
| acc-x-max | gyr-x-max | ori-angle-x-max |
| acc-x-var | gyr-x-var | ori-angle-x-var |
| acc-x-std | gyr-x-std | ori-angle-x-std |
| acc-y-min | gyr-y-min | ori-angle-y-min |
| acc-y-mean | gyr-y-mean | ori-angle-y-mean |
| acc-y-max | gyr-y-max | ori-angle-y-max |
| acc-y-var | gyr-y-var | ori-angle-y-var |
| acc-y-std | gyr-y-std | ori-angle-y-std |
| acc-z-min | gyr-z-min | ori-angle-z-min |
| acc-z-mean | gyr-z-mean | ori-angle-z-mean |
| acc-z-max | gyr-z-max | ori-angle-z-max |
| acc-z-var | gyr-z-var | ori-angle-z-var |
| acc-z-std | gyr-z-std | ori-angle-z-std |
| acc-sma | gyr-sma | mag-sma |
| acc-pcorr-xy | gyr-pcorr-xy | mag-pcorr-xy |
| acc-pcorr-xz | gyr-pcorr-xz | mag-pcorr-xz |
| acc-pcorr-yz | gyr-pcorr-yz | mag-pcorr-yz |
| acc-x-min-max-diff | - | - |
| acc-x-amprange | - | - |
| acc-x-mean-1quarter | - | - |
| acc-x-mean-2quarter | - | - |
| acc-x-mean-3quarter | - | - |
| acc-x-mean-4quarter | - | - |
| acc-x-aptd | gyr-x-aptd | - |
| acc-y-aptd | gyr-y-aptd | - |
| acc-z-aptd | gyr-z-aptd | - |
| acc-x-mad | gyr-x-mad | - |
| acc-y-mad | gyr-y-mad | - |
| acc-z-mad | gyr-z-mad | - |
| acc-x-energy | gyr-x-energy | mag-x-energy |
| acc-y-energy | gyr-y-energy | mag-y-energy |
| acc-z-energy | gyr-z-energy | mag-z-energy |
| acc-x-entropy | gyr-x-entropy | mag-x-entropy |
| acc-y-entropy | gyr-y-entropy | mag-y-entropy |
| acc-z-entropy | gyr-z-entropy | mag-z-entropy |
| acc-x-kurtosis | gyr-x-kurtosis | mag-x-kurtosis |
| acc-y-kurtosis | gyr-y-kurtosis | mag-y-kurtosis |
| acc-z-kurtosis | gyr-z-kurtosis | mag-z-kurtosis |
| activity-class | | |